\documentclass{article}
\usepackage{spconf,amsmath,graphicx}

\usepackage{url}

\usepackage{enumitem}
\setlist{nosep, leftmargin=14pt}

\usepackage{mwe} 

\usepackage{amsmath}
\usepackage{amssymb}
\usepackage{xcolor}
\usepackage{booktabs}
\usepackage{tikz}
\usetikzlibrary{positioning,shapes,arrows,fit,backgrounds,calc}
\usepackage{titlesec}
\usepackage[utf8]{inputenc}
\usetikzlibrary{arrows.meta,positioning,fit,calc,shapes.misc,shapes.geometric,decorations.pathreplacing}

\def\x{{\mathbf x}}

\title{INTERACT-CMIL: Multi-Task Shared Learning and Inter-Task Consistency for Conjunctival Melanocytic Intraepithelial Lesion Grading}
\name{%
\parbox{\linewidth}{Mert Ikinci$^{5}$\thanks{M. Ikinci has done the work during his Master's internship at Albarqouni Lab. at the University Hospital Bonn, Germany.} \qquad Luna Toma$^{2}$\thanks{M. Ikinci and L. Toma shared the first author position.} \qquad Karin U. Loeffler$^{2}$ \qquad Leticia Ussem$^{2}$ \qquad Daniela Süsskind$^{3}$ \\
\textit{Julia M. Weller$^{4}$ \qquad Yousef  Yeganeh$^{5,6}$ \qquad Martina C. Herwig-Carl$^{2}$ \qquad Shadi Albarqouni$^{1,5,7}$}\thanks{S.Albarqouni is the corresponding author: shadi.albarqouni@ukbonn.de}
}}
 \address{%
 \parbox{\linewidth}{\centering$^{1}$ Clinic for Diagnostic and Interventional Radiology, University Hospital Bonn, Germany \\
 $^{2}$ Department of Ophthalmology, Division of Ophthalmic Pathology, \\University Hospital Bonn, Germany\\
 $^{3}$ Centre of Ophthalmology, Eberhard Karls University of Tübingen \\and Comprehensive Cancer Center, University Hospital Tübingen, Germany\\
 $^{4}$ Department of Ophthalmology, Friedrich-Alexander University Erlangen-Nürnberg, Germany\\
 $^{5}$ TUM School of Computation, Information and Technology, Technical University of Munich, Germany\\
 $^{6}$ Munich Center for Machine Learning, Germany \\
 $^{7}$ Helmholtz AI, Helmholtz Center Munich, Germany
 }}

\begin{document}
%
\maketitle

\begin{abstract}
Accurate grading of Conjunctival Melanocytic Intraepithelial Lesions (CMIL) is essential for treatment and melanoma prediction but remains difficult due to subtle morphological cues and interrelated diagnostic criteria. We introduce \textbf{INTERACT-CMIL}, a multi-head deep learning framework that jointly predicts five histopathological axes; WHO4, WHO5, horizontal spread, vertical spread, and cytologic atypia, through \emph{Shared Feature Learning with Combinatorial Partial Supervision} and an \emph{Inter-Dependence Loss} enforcing cross-task consistency. Trained and evaluated on a newly curated, multi-center dataset of 486 expert-annotated conjunctival biopsy patches from three university hospitals, INTERACT-CMIL achieves consistent improvements over CNN and foundation-model (FM) baselines, with relative macro F\textsubscript{1} gains up to 55.1\% (WHO4) and 25.0\% (vertical spread). The framework provides coherent, interpretable multi-criteria predictions aligned with expert grading, offering a reproducible computational benchmark for CMIL diagnosis and a step toward standardized digital ocular pathology.
\end{abstract}


\section{Introduction}
\label{sec:intro}

Conjunctival Melanocytic Intraepithelial Lesions (CMIL) are recognized precursors of conjunctival melanoma, a rare yet potentially lethal ocular malignancy. Timely and accurate classification of CMIL is critical, as diagnostic uncertainty can substantially affect patient management and outcomes~\cite{bresler2022conjunctival}. Current diagnostic practice relies on histopathological examination of Hematoxylin and Eosin (H\&E)-stained sections and immunohistochemical markers. However, these assessments remain subjective, exhibiting significant inter-observer variability, particularly for low-grade lesions~\cite{milman2021validation,mudhar2024multicenter}. This variability underscores the urgent need for objective, reproducible, and interpretable computational tools to support diagnostic decision-making.

CMIL grading has evolved from the earlier \textit{primary acquired melanosis (PAM)} classification to the Conjunctival Melanocytic Intraepithelial Neoplasia (C-MIN) scoring system proposed by Damato and Coupland~\cite{damato2009management}, and more recently, to the World Health Organization’s (WHO) 4th and 5th edition tumor classification frameworks~\cite{grossniklaus2018classification,milman20235th}. These systems incorporate key histologic criteria such as  growth pattern and melanocytic atypia, highlighting the necessity of multi-class and multi-criteria diagnostic modeling. Translating these complex grading schemes into computational pathology frameworks requires models that balance predictive accuracy with interpretability and clinical alignment.

\begin{figure}[!t]
    \centering
    \includegraphics[width=0.48\linewidth]{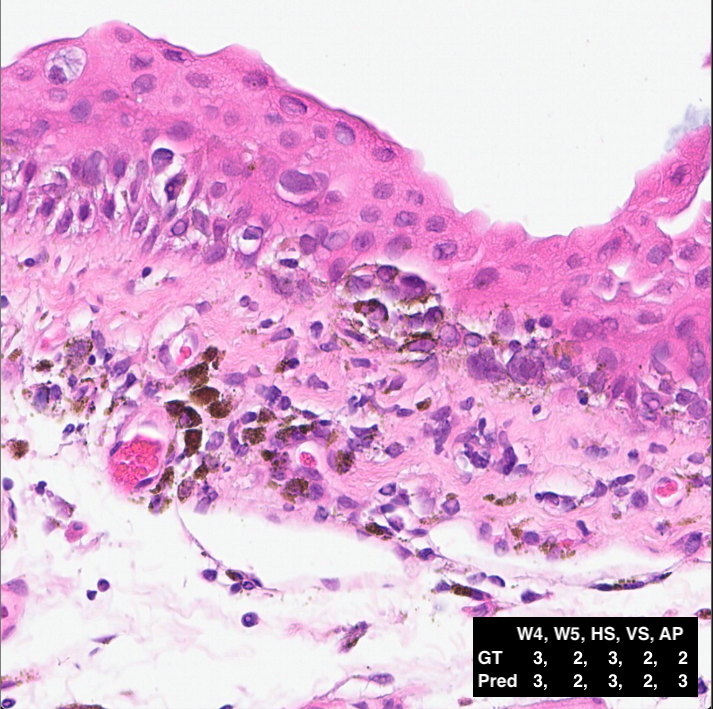}
    \includegraphics[width=0.48\linewidth]{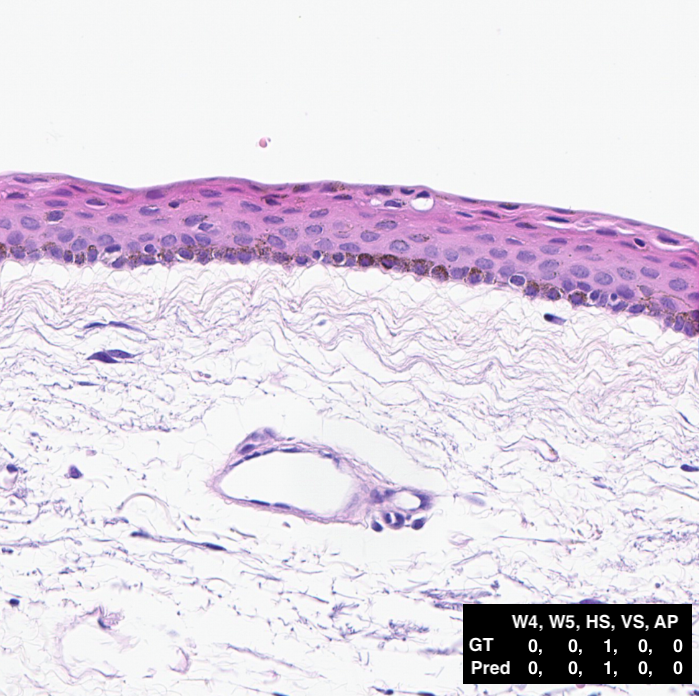}
    \caption{Representative conjunctival biopsy patches illustrating the diagnostic complexity of CMIL grading.  Each image is annotated with the five grading criteria: WHO4 (W4), WHO5 (W5), horizontal spread (HS), vertical spread (VS), and cytologic atypia (AP), showing the expert ground truth (GT) and model prediction (Pred.).   \textbf{Left:} a high-grade lesion exhibiting nests of  melanocytes and marked melanocytic atypia, accurately predicted by our method across all criteria.  \textbf{Right:} a benign lesion with limited melanocytic proliferation and orderly epithelium, likewise correctly classified.}
    \label{fig:golden}
\end{figure}

Recent advances in deep learning, particularly foundation models trained on large-scale pathology datasets~\cite{van2021deep,campanella2019clinical,chen2024towards,wang2024pathology}, have substantially improved performance in tissue classification, biomarker identification, and prognostic prediction. Despite these developments, CMIL remains largely unexplored due to its rarity, limited data availability, and the complexity of its histopathologic taxonomy.

To address these challenges, we introduce INTERACT-CMIL, a multi-head deep learning framework built upon the CHIEF foundation model~\cite{wang2024pathology}, a large-scale pathology encoder pretrained on diverse whole-slide images to capture transferable histomorphologic representations, for robust and interpretable CMIL classification. The proposed architecture employs five task-specific classification heads corresponding to major histological criteria. During training, a combinatorial supervision strategy selectively activates three heads per iteration to enhance regularization and mitigate overfitting. Furthermore, an \textit{inter-dependence loss} explicitly models correlations among grading criteria, fostering consistency across related diagnostic tasks and aligning predictions with established WHO and C-MIN standards. Rather than relying on random sampling, the strategy systematically cycles through all possible three-head combinations, ensuring comprehensive coverage of the task space and consistent contribution from every diagnostic head. 

\noindent Our \textbf{contributions} are summarized as:
\begin{itemize}
    \item We present the first deep learning framework specifically designed for CMIL classification, integrating WHO and C-MIN grading paradigms.
    \item We introduce a novel multi-head architecture with combinatorial supervision to promote robustness and balanced task learning.
    \item We propose an inter-dependence loss that captures cross-criterion relationships, enhancing model interpretability and diagnostic coherence.
\end{itemize}

\section{Methodology}
\label{sec:method}

\tikzset{
  >={Latex[length=2.2mm]},
  box/.style={draw, rounded corners=2pt, thick, align=center, fill=gray!5},
  mod/.style={box, fill=blue!8},
  head/.style={box, fill=purple!10},
  loss/.style={box, fill=orange!15},
  tensor/.style={box, fill=teal!8},
  frozen/.style={box, fill=cyan!10, dashed},
  tinylabel/.style={font=\scriptsize, inner sep=1pt},
  arrow/.style={->, line width=0.9pt},
  dparrow/.style={->, line width=0.9pt, dashed},
  note/.style={font=\footnotesize, align=center}
}

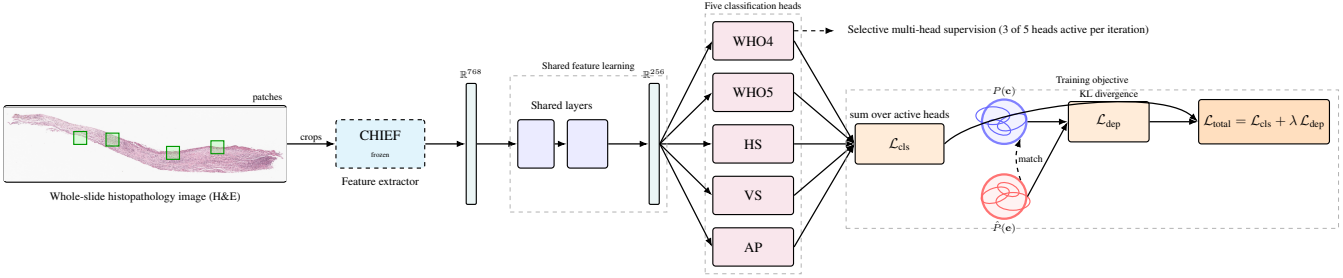
\begin{figure*}[t]
\centering
\resizebox{\textwidth}{!}{%
\begin{tikzpicture}[node distance=8mm and 10mm]

\node[head, minimum width=2.0cm, minimum height=0.95cm] (h3) at (0,0) {HS};
\node[head, above=3mm of h3, minimum width=2.0cm, minimum height=0.95cm] (h2) {WHO5};
\node[head, above=3mm of h2, minimum width=2.0cm, minimum height=0.95cm] (h1) {WHO4};
\node[head, below=3mm of h3, minimum width=2.0cm, minimum height=0.95cm] (h4) {VS};
\node[head, below=3mm of h4, minimum width=2.0cm, minimum height=0.95cm] (h5) {AP};

\node[draw=gray!60, dashed, fit=(h1) (h5), inner sep=5pt] (headsbox) {};
\node[tinylabel, anchor=south, yshift=2pt] at (headsbox.north) {Five classification heads};

\node[draw, rounded corners=1pt, thick, fill=teal!8, left=13mm of h3,
      minimum width=0.25cm, minimum height=3.0cm, inner sep=2pt] (feat256bar) {};
\node[tinylabel, anchor=south, yshift=2pt] at (feat256bar.north) { $\mathbb{R}^{256}$};

\foreach \h in {h1,h2,h3,h4,h5} {
  \draw[arrow] (feat256bar.east) -- (\h.west);
}

\node[mod, left=10mm of feat256bar, minimum width=1.0cm, minimum height=1.2cm] (shared2) {};
\node[mod, left=3mm of shared2, minimum width=0.9cm, minimum height=1.2cm] (shared1) {};
\node[note, anchor=south, yshift=2pt] at ($(shared1.north)!0.5!(shared2.north)$) {Shared layers};
\draw[arrow] (shared1.east) -- (shared2.west);
\draw[arrow] (shared2.east) -- (feat256bar.west);

\node[draw, rounded corners=1pt, thick, fill=teal!8, left=10mm of shared1,
      minimum width=0.25cm, minimum height=3.0cm, inner sep=2pt] (feat768bar) {};
\node[tinylabel, anchor=south, yshift=2pt] at (feat768bar.north) {$\mathbb{R}^{768}$};
\draw[arrow] (feat768bar.east) -- (shared1.west);

\node[frozen, left=10mm of feat768bar, minimum width=2.2cm, minimum height=1.2cm] (chief) {CHIEF\\\tiny frozen};
\node[note, below=1mm of chief] {Feature extractor};
\draw[arrow] (chief.east) -- (feat768bar.west);

\node[box, left=12mm of chief, inner sep=0pt, minimum width=7.0cm, minimum height=1.9cm] (wsi) {\includegraphics[width=7.0cm]{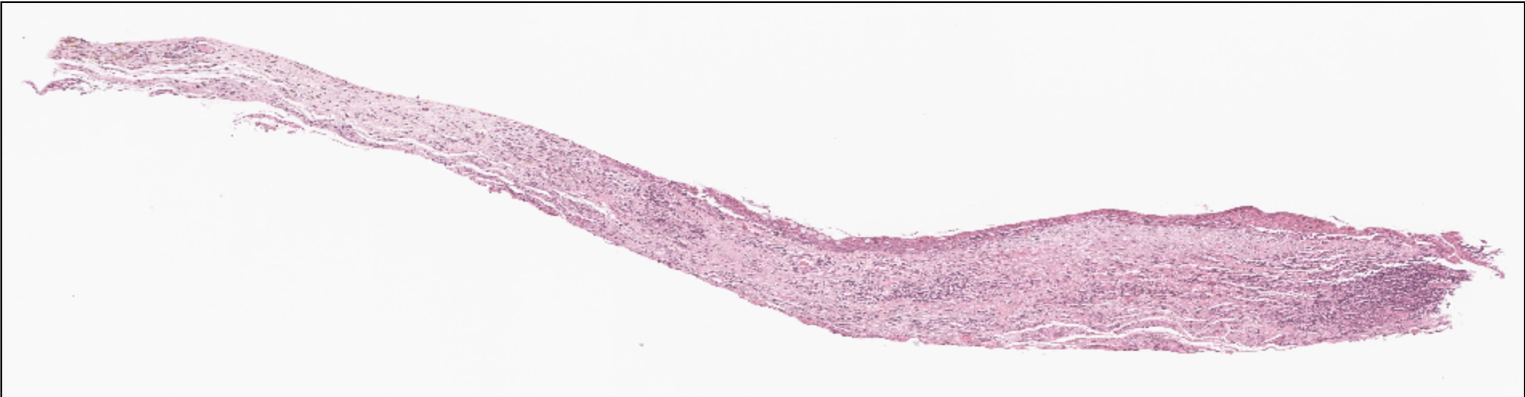}};
\node[note, below=1mm of wsi] {Whole-slide histopathology image (H\&E)};

\foreach \x/\y in {0.9/0.72, 1.7/0.68, 3.2/0.35, 4.3/0.49} {
  \draw[thick, draw=green!60!black, fill=green!50, fill opacity=0.25, line width=0.8pt]
    ($(wsi.south west)!0.12!(wsi.north east) + (\x, \y)$)
    rectangle ++(0.32,0.32);
}

\node[tinylabel, anchor=south east, xshift=-2pt, yshift=1pt] at (wsi.north east) {patches};

\draw[arrow] (wsi.east) -- node[above, tinylabel]{crops} (chief.west);

\node[draw=gray!60, dashed, fit=(shared1) (shared2) (feat256bar), inner sep=5pt] (sharedbox) {};
\node[tinylabel, anchor=south, yshift=2pt] at (sharedbox.north) {Shared feature learning};

\node[loss, right=15mm of h3, minimum width=2.2cm, minimum height=1.0cm] (lcls) {$\mathcal{L}_{\text{cls}}$};
\node[note, above=0mm of lcls] {sum over active heads};
\foreach \h in {h1,h2,h3,h4,h5} { \draw[arrow] (\h.east) -- (lcls.west); }

\node[draw, circle, minimum width=1.1cm, ultra thick, draw=blue!50, fill=blue!5, right=9mm of lcls, yshift=5.5mm] (priorA) {};
\draw[thick, draw=blue!60] ($(priorA.center)+(-0.28,0.15)$) ellipse (0.42 and 0.22);
\draw[thick, draw=blue!60, rotate around={20:(priorA.center)}] ($(priorA.center)+(0.18,-0.08)$) ellipse (0.38 and 0.18);
\draw[thick, draw=blue!60, rotate around={-25:(priorA.center)}] ($(priorA.center)+(-0.05,-0.18)$) ellipse (0.30 and 0.16);
\node[tinylabel, above=0mm of priorA] {$P(\mathbf{c})$};

\node[draw, circle, minimum width=1.1cm, ultra thick, draw=red!55, fill=red!5, below=7mm of priorA] (postA) {};
\draw[thick, draw=red!60] ($(postA.center)+(0.22,-0.12)$) ellipse (0.42 and 0.22);
\draw[thick, draw=red!60, rotate around={18:(postA.center)}] ($(postA.center)+(-0.20,0.10)$) ellipse (0.36 and 0.18);
\draw[thick, draw=red!60, rotate around={-22:(postA.center)}] ($(postA.center)+(0.03,0.20)$) ellipse (0.28 and 0.15);
\node[tinylabel, below=0mm of postA] {$\hat{P}(\mathbf{c})$};

\draw[dparrow] (postA.north east) to[bend left=20] node[right, tinylabel]{match} (priorA.south east);

\node[loss, right=10mm of priorA, minimum width=2.0cm, minimum height=1.0cm] (ldep) {$\mathcal{L}_{\text{dep}}$};
\node[tinylabel, above=0mm of ldep] {KL divergence};
\draw[arrow] (priorA.east) -- (ldep.west);
\draw[arrow] (postA.east) -- (ldep.west);

\node[loss, right=12mm of ldep, minimum width=2.8cm, minimum height=1.1cm, fill=orange!25] (ltotal) {$\mathcal{L}_{\text{total}}=\mathcal{L}_{\text{cls}}+\lambda\,\mathcal{L}_{\text{dep}}$};

\draw[arrow] (lcls.east) .. controls +(0.9,0.9) and +(-1.0,0.9) .. (ltotal.west);
\draw[arrow] (ldep.east) -- (ltotal.west);

\node[draw=gray!60, dashed, fit=(lcls) (priorA) (postA) (ldep) (ltotal), inner sep=6pt] (lossbox) {};
\node[tinylabel, anchor=south, yshift=2pt] at (lossbox.north) {Training objective};

\draw[dparrow, thick] ($(h1.east)+(0,0.25)$) -- ++(1.0,0)
  node[note, right=2mm] {Selective multi-head supervision (3 of 5 heads active per iteration)};

\end{tikzpicture}
}
\caption{Pipeline overview. Multi-center data (left) produce whole-slide images (WSI). Patches are extracted and fed to a \emph{frozen} CHIEF feature extractor yielding a 768-D vector, transformed by shared layers into a 256-D shared representation. Five heads (\textbf{WHO4}, \textbf{WHO5}, \textbf{HS}, \textbf{VS}, \textbf{AP}) produce task probabilities; at each iteration, three heads are updated (selective supervision). Training minimizes \(\mathcal{L}_{\text{total}}=\mathcal{L}_{\text{cls}}+\lambda\,\mathcal{L}_{\text{dep}}\), where \(\mathcal{L}_{\text{dep}}\) aligns the predicted joint distribution \(\hat P\) with the empirical prior \(P\) (schematized as 2D Gaussians).}
\label{fig:main_arch}
\end{figure*}

The histopathological assessment of CMIL depends on multiple interrelated diagnostic criteria that jointly determine lesion grade and malignant potential. Specifically, five key axes: (1) WHO4 classification, (2) WHO5 classification, (3)  horizontal spread/growth pattern (HS), (4) vertical spread (VS), and (5) cellular atypia (3-5 belong to the C-MIN classification)—are used to characterize disease severity. These criteria exhibit substantial interdependence: for instance, severe atypia often coincides with increased vertical proliferation, while the WHO5 classification partially subsumes features of WHO4. Modeling these axes independently disregards such relationships and risks inconsistent or clinically implausible predictions.

To address this, we propose, INTERACT-CMIL, a unified multi-head architecture, termed \emph{Shared Feature Learning with Combinatorial Partial Supervision (SFCS)}, designed to capture both the distinct semantics and the latent dependencies across these five diagnostic tasks. The framework operates on a shared feature representation learned from histopathological patches and employs selective multi-head training to encourage robust, interpretable, and dependency-aware learning as illustrated in Fig~\ref{fig:main_arch}, which outlines the core workflow of our approach.

\subsection{Feature Representation}

Given a dataset \(\mathcal{D} = \{I_1, I_2, \ldots, I_N\}\) of \(N\) histopathological patches \(I \in \mathbb{R}^{h \times w \times c}\), each patch is embedded into a feature vector \(\boldsymbol{x} = f_{\text{enc}}(I)\), where \(f_{\text{enc}}\) denotes a pretrained feature extractor. The feature vector is projected by a shared transformation network \(f_{\text{shared}}\) to obtain a latent representation \(z = f_{\text{shared}}(\boldsymbol{x}) \in \mathbb{R}^{h}\), capturing shared morphological attributes relevant to all diagnostic tasks.

\subsection{Selective Multi-Head Supervision}

The model comprises \(T=5\) classification heads \(\{g_t\}_{t=1}^{5}\), each corresponding to one of the five diagnostic axes described above. For a given input \(z\), each head produces a probability vector over its class set:
$
\tilde{p}_t = \text{softmax}(g_t(z)).
$

Instead of jointly optimizing all heads, SFCS adopts a \emph{combinatorial partial supervision} strategy: during each training iteration, only a subset of heads \(\mathcal{T} \subset \{1, \ldots, 5\}\) of size \(|\mathcal{T}|=3\) is activated. Across training, all possible three-head combinations are sampled, ensuring uniform exposure to diverse supervision contexts. This selective optimization mitigates overfitting and encourages the shared representation to encode features that generalize across correlated but non-identical tasks.
The classification objective for the active subset \(\mathcal{T}\) is expressed as:
\[
\mathcal{L}_{\text{cls}} = 
\sum_{t \in \mathcal{T}} 
\Big[- \sum_{j=1}^{k_t} y_{t,j} \log \tilde{p}_{t,j}\Big],
\]
where \(y_t\) denotes the one-hot ground-truth label vector for task \(t\) with \(k_t\) classes.

\subsection{Inter-Task Dependency Regularization}

Since CMIL grading axes exhibit structured correlations, enforcing consistency across them is crucial. We introduce a divergence-based dependency regularizer that aligns the predicted joint task distributions with the empirical co-occurrence statistics observed in the data. For the current task subset \(\mathcal{T} = \{t_1, t_2, t_3\}\), the predicted and empirical joint distributions are estimated as:
\[
\hat{P} = \frac{1}{B} \sum_{i=1}^{B} 
\tilde{p}^{(i)}_{t_1} \otimes \tilde{p}^{(i)}_{t_2} \otimes \tilde{p}^{(i)}_{t_3}, \quad
P = \frac{1}{B} \sum_{i=1}^{B} 
y^{(i)}_{t_1} \otimes y^{(i)}_{t_2} \otimes y^{(i)}_{t_3},
\]
where $B$ denotes the batch size and $\otimes$ represents the outer product operator. The dependency loss is then defined as:
\[
\mathcal{L}_{\text{dep}} =
\sum_{\mathbf{c} \in \mathcal{C}} 
P(\mathbf{c}) \log \frac{P(\mathbf{c})}{\hat{P}(\mathbf{c}) + \epsilon},
\]
with \(\mathcal{C}\) denoting all possible label combinations and $\epsilon > 0$ a small constant ensuring numerical stability.

\subsection{Overall Objective}

The final objective integrates task-level accuracy and dependency consistency:
\[
\mathcal{L}_{\text{total}} =
\mathcal{L}_{\text{cls}}
+ \lambda \, \mathcal{L}_{\text{dep}},
\]
where \(\lambda\) controls the influence of dependency regularization. Only the active task heads and shared layers are updated in each iteration to maintain disentangled optimization across heads. Overall, the proposed framework introduces three key methodological advances:  
(i) a multi-head formulation explicitly aligned with the five clinically defined CMIL grading axes;  
(ii) a combinatorial partial supervision scheme that promotes robust shared representation learning across correlated tasks; and  
(iii) a divergence-based dependency loss that enforces semantic coherence among task outputs, yielding interpretable and clinically consistent predictions.

\section{Experiments and Results}
\label{sec:experiments}

We evaluate the proposed framework on a held-out test set of conjunctival biopsy patches with expert-consensus labels for the five CMIL axes: WHO4, WHO5, horizontal spread (HS), vertical spread (VS), and atypia (cf.~Sec.~\ref{sec:method}). Train/validation/test splits are patient-disjoint. All methods follow the same preprocessing and produce a single prediction per task per image. 

\subsection{Experimental setup.}

\subsubsection{Dataset}
Our dataset comprises high-resolution histopathology crops from conjunctival biopsies collected across three German university hospitals (Tübingen, Bonn, and Erlangen). Each crop was independently reviewed and jointly adjudicated by specialist ophthalmic pathologists to ensure expert-consensus labeling and reduce observer variability.  
Every sample is annotated along five clinically relevant axes: WHO4 and WHO5 grades, and three C-MIN components: HS, VS, and AP. In total, the dataset contains 486 annotated crops, with Bonn contributing the largest share (333), followed by Erlangen (135) and Tübingen (18), introducing beneficial heterogeneity in staining and acquisition.  
Label distributions reveal a predominance of higher-grade lesions (WHO5=2, WHO4=3), with C-MIN scores spanning 0–10 and peaking around 6–8, reflecting a clinically meaningful range from low to severe atypia. Analysis of inter-feature relationships confirms strong dependencies between vertical spread and atypia (300 samples with zero discrepancy), consistent with established pathological correlations.  
This expert-curated, multi-center dataset thus provides a robust foundation for developing multi-task learning models.

\subsubsection{Baselines}
We compare against: (i) \textbf{BaseCNN}, a ResNet-18 backbone pretrained on ImageNet, trained end-to-end with five heads jointly optimized using the sum of task losses; and (ii) \textbf{BaseCHIEF}, a single-task setup using frozen pathology embeddings from the CHIEF foundation model, which is pretrained on large-scale whole-slide images to capture transferable histomorphologic representations. Our method, \textbf{INTERACT-CMIL}, employs selective multi-head supervision and a dependency regularizer.

\subsubsection{Evaluation Metrics}
\begin{figure*}[!ht]
    \centering
    \includegraphics[width=0.195\linewidth, height=0.15\textheight]{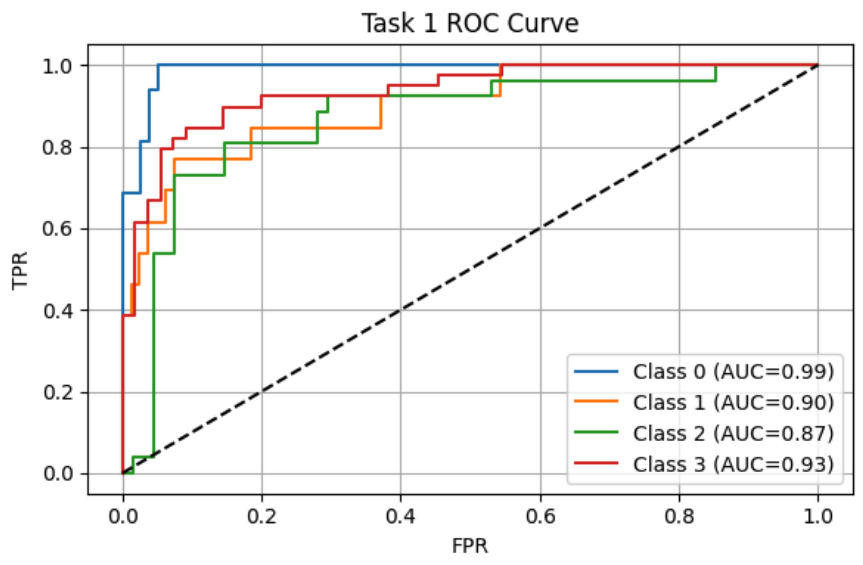}
    \includegraphics[width=0.195\linewidth, height=0.15\textheight]{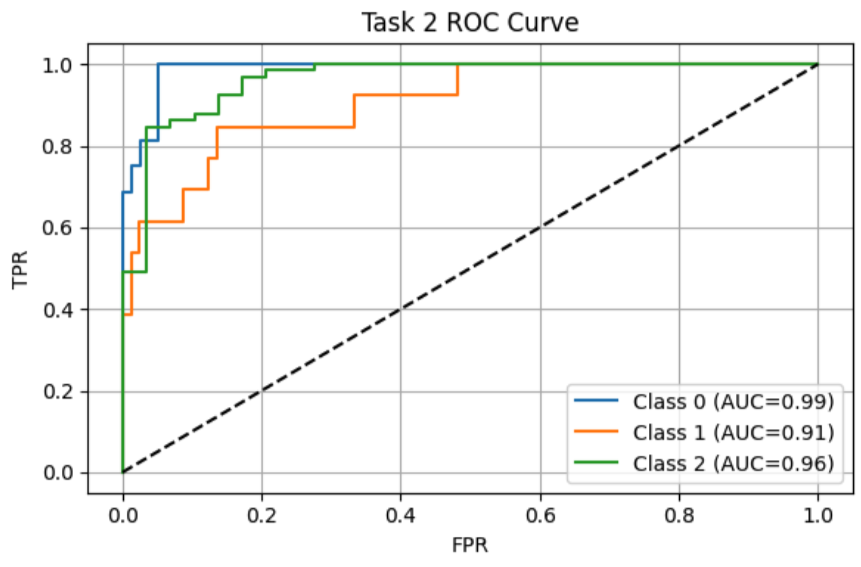}
    \includegraphics[width=0.195\linewidth, height=0.15\textheight]{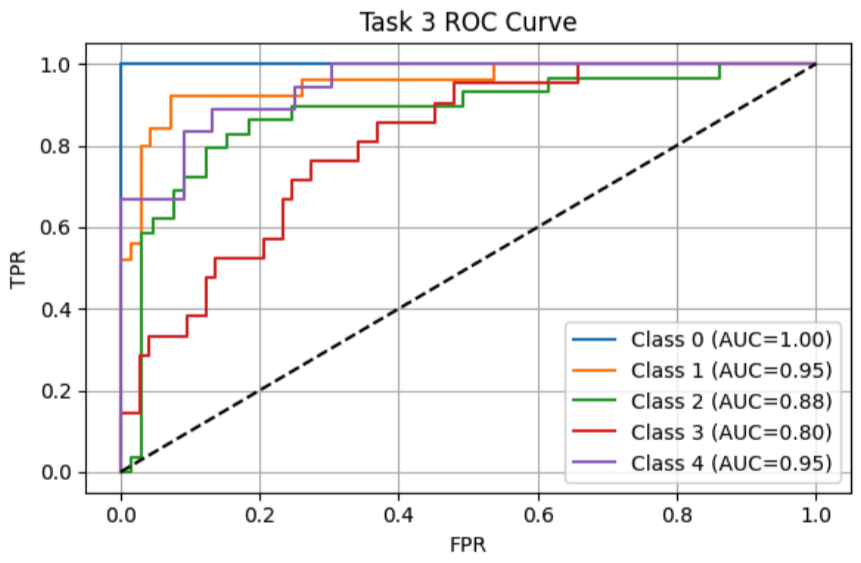}
    \includegraphics[width=0.195\linewidth, height=0.15\textheight]{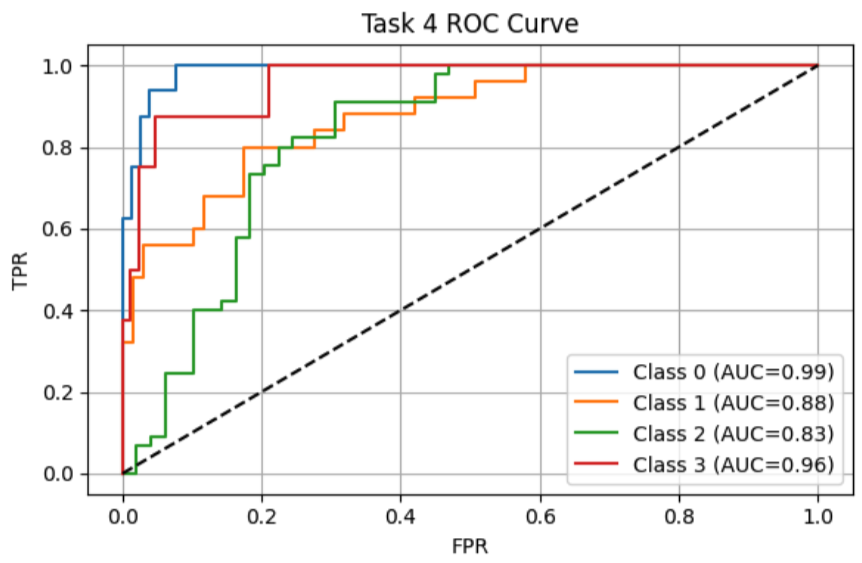}
    \includegraphics[width=0.195\linewidth, height=0.15\textheight]{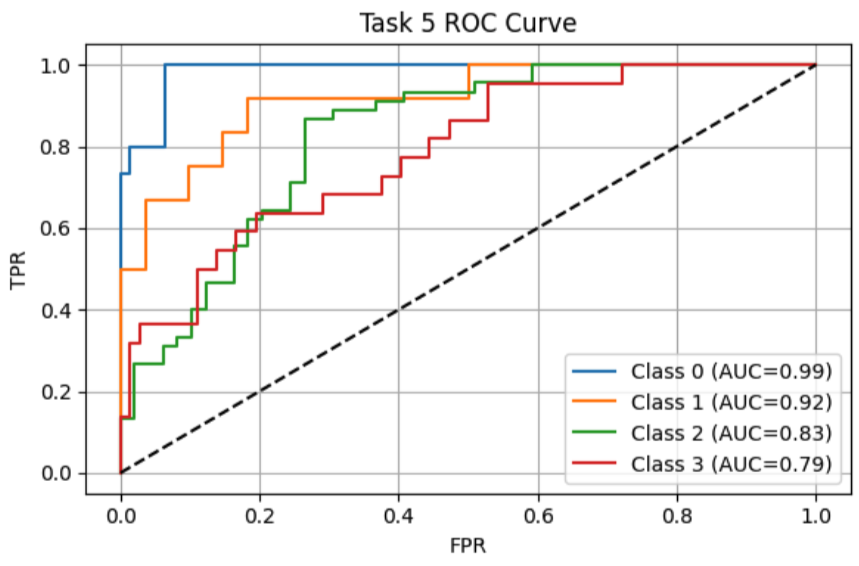}
    \caption{ROC Curves of our method for different tasks; WHO4, WHO5, Horizontal Spread, Vertical Spread, and Atypia, respectively.}
    \label{fig:roc_12}
\end{figure*}

Our primary evaluation metric is \emph{macro} F\textsubscript{1} per task, reported as mean (median)~$\pm$~standard deviation over 5-fold cross-validation (Tab.~\ref{tab:baseline_comparison}). Macro F\textsubscript{1} is computed per-task across 3 classes (for WHO5),  4 for WHO4,  4 for HS, 3 for VS, and 3 for AP. Macro F\textsubscript{1} treats all classes equally and is preferred under class imbalance. For completeness, ROC curves and AUC are provided in Fig.~\ref{fig:roc_12} as secondary, but the comparisons and all numbers are reported in Tab.~\ref{tab:baseline_comparison}. Implementation was performed in PyTorch; training converged within ~100 epochs on a single NVIDIA GPU.

\begin{table*}[!t]
    \centering
    \caption{Comparison of baseline models against the proposed INTERACT-CMIL framework and the ablation of the Inter-Dependence Loss (dep.), Temperature Scaling (temp.), and the Selective Multi-Head Supervision (sel.) across five CMIL classification tasks is reported as Mean (Median) ± Standard deviation over 5-fold
    cross-validation.}
    \label{tab:baseline_comparison}
    \resizebox{\textwidth}{!}{
    \begin{tabular}{l|c|c|c|c|c}
        \toprule
        \textbf{Model} & \textbf{WHO4} & \textbf{WHO5} & \textbf{Horizontal Spread} & \textbf{Vertical Spread} & \textbf{Atypia} \\
        \midrule
        BaseCNN & 0.4293 (0.4515) $\pm$ 0.0331 & 0.5848 (0.6300) $\pm$ 0.1344 & 0.3434 (0.3378) $\pm$ 0.0368 & 0.2902 (0.2874) $\pm$ 0.0144 & 0.3687 (0.3682) $\pm$ 0.0431 \\
        BaseCHIEF & 0.4911 (0.4949) $\pm$ 0.0107 & 0.7029 (0.6988) $\pm$ 0.0105 & 0.6134 (0.6105) $\pm$ 0.0273 & 0.6091 (0.6162) $\pm$ 0.0242 & 0.6253 (0.6220) $\pm$ 0.0228 \\
        \midrule
        INTERACT-CMIL (Ours) &  \textbf{0.7617 (0.7632) $\pm$ 0.0331} & \textbf{0.8766 (0.8729) $\pm$ 0.0117} & \textbf{0.7066 (0.7028) $\pm$ 0.0268} & \textbf{0.7613 (0.7565) $\pm$ 0.0216} & \textbf{0.7032 (0.7034) $\pm$ 0.0323} \\
        \quad -- dep.  &   0.7397 (0.7536) $\pm$ 0.0319 & 0.8613 (0.8603) $\pm$ 0.0148 & 0.7045 (0.7076) $\pm$ 0.0207 & 0.7542 (0.7559) $\pm$ 0.0183 & 0.6869 (0.6716) $\pm$ 0.0347 \\
        \quad -- dep. -- temp. & 0.6924 (0.6938) $\pm$ 0.0257  & 0.8172 (0.8332) $\pm$ 0.0208 & 0.6879 (0.6881) $\pm$ 0.0238 & 0.7103 (0.7113) $\pm$ 0.0178 & 0.6461 (0.6516) $\pm$ 0.0272 \\
        \quad -- dep. -- temp. -- sel. & 0.6120 (0.6175) $\pm$ 0.0340 & 0.7957 (0.7955) $\pm$ 0.0142 & 0.6672 (0.6596) $\pm$ 0.0380 & 0.6602 (0.6582) $\pm$ 0.0156 & 0.6328 (0.6411) $\pm$ 0.0142 \\
        \bottomrule
    \end{tabular}
    }
\end{table*}

\subsection{Main Results}
Table~\ref{tab:baseline_comparison} summarizes performance across the five tasks. Our proposed method, namely \textbf{INTERACT-CMIL}, attains the best results on all axes, with consistent gains over both single-task and conventional multi-task baselines. Our method achieves a mean macro F\textsubscript{1} on all five CMIL axes:
WHO4 \textbf{0.7617},
WHO5 \textbf{0.8766},
HS \textbf{0.7066},
VS \textbf{0.7613},
and Atypia \textbf{0.7032}.
Relative to the strongest single-task baseline (\textbf{BaseCHIEF}), our method improves macro F\textsubscript{1} by
{55.10\%} (WHO4),
{24.71\%} (WHO5),
{15.19\%} (HS),
{24.99\%} (VS),
and {12.46\%} (Atypia);
gains over \textbf{BaseCNN} are larger across all tasks. 
The results indicate that selective multi-head supervision with dependency regularization yields more consistent improvements than direct feature cross-attention.
Qualitative comparisons in Fig.~\ref{fig:golden} indicate close agreement between ground-truth (GT) and predictions (Pred) across different tasks, including a borderline case where human variability is quite high.

\subsection{Ablations}
To evaluate the contribution of each component in INTERACT-CMIL, we progressively remove the inter-dependence loss (dep.), temperature scaling (temp.), and selective multi-head supervision (sel.), as summarized in Table~\ref{tab:baseline_comparison}. Each component provides additive gains across all five diagnostic tasks.

\subsubsection{Pretrained Features vs. Scratch}
Using pathology-pretrained features (BaseCHIEF) provides a substantial boost over a Resnet-18 CNN trained from scratch (BaseCNN), with mean macro F1 increasing by over 20\% across tasks. This confirms that foundation model representations are crucial for small, heterogeneous CMIL datasets.

\subsubsection{Dependency Regularization}
Removing the inter-dependence loss (--dep.) causes the largest single drop in performance (mean macro F1 decrease $\sim$2--3\%), particularly on atypia and vertical spread (–1.6 and –0.7 absolute). This shows that modeling task correlations improves label coherence and high-grade lesion detection.

\subsubsection{Temperature Scaling}
Disabling both dependency and temperature scaling (--dep.--temp.) yields a further decline of $\sim$5--6\% across tasks (e.g., WHO5: 0.8613 $\rightarrow$ 0.8172, VS: 0.7542 $\rightarrow$ 0.7103). This demonstrates that temperature calibration sharpens inter-task decision boundaries and stabilizes predictions.

\subsubsection{Selective Multi-Head Supervision}
Eliminating selective supervision (--dep.--temp.--sel.) leads to the most significant cumulative drop, averaging over 8\% below the full model (e.g., WHO4: 0.7617 $\rightarrow$ 0.6120). Selective head updates thus prevent gradient domination, improving balance between easier (HS) and harder (VS, AP) tasks and acting as a strong regularizer.

Overall, dependency modeling contributes $\sim$3\% of the total improvement, temperature scaling another $\sim$4\%, and selective supervision an additional $\sim$4\%, jointly explaining the $\sim$10\% aggregate gain of INTERACT-CMIL over the strongest baseline.

\section{Discussion and Conclusion}
\label{sec:conclusion}

INTERACT-CMIL demonstrates that structured multi-task learning can effectively capture the intertwined histopathological criteria that define CMIL grading. By combining shared feature learning, selective supervision, and inter-task dependency regularization, the model produces coherent, clinically consistent predictions that align with expert consensus and the established WHO/C-MIN frameworks. 

Quantitatively, INTERACT-CMIL achieves mean macro F\textsubscript{1} scores of 0.76 (WHO4), 0.88 (WHO5), 0.71 (horizontal spread), 0.76 (vertical spread), and 0.70 (atypia), surpassing the strongest baseline (BaseCHIEF) by relative margins of +55.1\%, +24.7\%, +15.2\%, +25.0\%, and +12.5\%, respectively. The largest gains occur for \textbf{vertical spread} and \textbf{cytologic atypia}, which rely on subtle morphological cues and inter-criterion consistency; areas where single-task or CNN-based models tend to fail. Performance on \textbf{horizontal spread} remains high and near-saturated, reflecting that this visually evident feature benefits less from multi-task context. These consistent cross-task improvements underscore the benefit of explicitly modeling diagnostic dependencies in multi-criteria pathology tasks.

The curated in-house dataset, consisting of 486 expert-annotated biopsy crops from three institutions, represents the first multi-center benchmark for computational CMIL grading. Its cross-site diversity in staining and acquisition conditions promotes generalization and provides a realistic foundation for studying inter-observer and inter-institutional variability.

Future work will extend this dataset, incorporate whole-slide representations, and explore domain adaptation and self-supervised pretraining to enhance robustness under clinical variability. Overall, INTERACT-CMIL establishes a reproducible, interpretable, and clinically aligned framework for automated CMIL classification, advancing computational support for ocular pathology diagnostics.


\noindent \textbf{Compliance with ethical standards.} 
All procedures involving human participants were conducted in accordance with the ethical standards of the institutional and/or national research committees and with the 1964 Declaration of Helsinki and its later amendments.
The project was approved by the Ethics Committee of the University of Bonn (328/16, 2025-340-BO).

\section*{Acknowledgments}
\label{sec:acknowledgments}
Y.Y. is partly funded by the Munich Center for Machine Learning.

\bibliographystyle{IEEEbib}
\bibliography{refs}

\end{document}